\documentclass[letterpaper, 10 pt, conference]{ieeeconf}  

\IEEEoverridecommandlockouts                              

\usepackage{times}
\usepackage{cite}
\usepackage{multicol}
\usepackage{multirow}
\usepackage[bookmarks=true]{hyperref}
\usepackage{tabularx}
\usepackage{mathtools}
\usepackage{subcaption}
\usepackage{booktabs}
\usepackage{wrapfig}
\usepackage[nameinlink,capitalise]{cleveref}  % For use of \cref
\let\labelindent\relax
\usepackage{enumitem}

\usepackage[most]{tcolorbox}

\usepackage{xcolor}
\definecolor{smerald}{RGB}{8, 104, 120}

\usepackage{cuted}
\usepackage{amsmath} % assumes amsmath package installed
\usepackage{amssymb}  % assumes amsmath package installed
\usepackage{caption}
\usepackage{adjustbox}

\graphicspath{{imgs}}

\usepackage{svg}

\usepackage{xcolor}

\usepackage{relsize}                        % Adjust sizes of math symbols
\usepackage{tensor}                         % Tensor index notation
\usepackage{ifthen}                         % Logic-based conditional commands
\usepackage[all,cmtip]{xy}                  % Diagrams and arrows
\usepackage{graphicx}                       % Include graphics
\usepackage{amsmath}
\usepackage{amssymb}
\usepackage{mathtools}

\usepackage{amsthm}

\usepackage{amsmath,amsfonts,bm}

\def\1{\bm{1}}

\def\rvx{{\mathbf{x}}}
\def\rvy{{\mathbf{y}}}

\def\vtheta{{\bm{\theta}}}

\def\va{{\bm{a}}}

\def\vo{{\bm{o}}}

\def\vs{{\bm{s}}}

\def\vv{{\bm{v}}}

\def\vx{{\bm{x}}}
\def\vy{{\bm{y}}}

\def\vsX{{\mathcal{X}}}
\def\vsY{{\mathcal{Y}}}

\DeclareMathAlphabet{\mathsfit}{\encodingdefault}{\sfdefault}{m}{sl}
\SetMathAlphabet{\mathsfit}{bold}{\encodingdefault}{\sfdefault}{bx}{n}

\newcommand{\operator}[1]{\mathsf{#1}}

\def\oE{{\operator{E}}}

\newcommand{\E}{\mathbb{E}}

\newcommand{\R}{\mathbb{R}}

\newcommand{\KL}{D_{\mathrm{KL}}}

\DeclareMathOperator*{\argmin}{arg\,min}

\newcommand{\CyclicGroup}[1][]{\mathbb{C}_{#1}}                             % Cyclic Group
\newcommand{\SE}[1][\dimEvolution]{\mathbb{SE}(#1)}                        % Special Euclidean Group
\newcommand{\G}[1][]{\mathbb{G}_{\scalebox{0.6}{$#1$}}}                     % Symmetry Group
\newcommand{\g}{g}                                                          % Group member
\newcommand{\Glact}[1][]
{
	\,{
			\mathrel{
				\mathsmaller{
					\triangleright_{\scalebox{0.65}{$#1$}}
				}
			}
		}\,
}                                                                            % Left group action    g \Glact x
\newcommand{\Gract}[1][]{\mathrel{\mathsmaller{\triangleleft}}}              % Right group action   x \Gract g
\newcommand{\Gconj}[1][]{\mathrel{\mathsmaller{\diamond}}}                   % Conjugate group action  g \Gconj := g \Glact X  \Gract \g^{-1}
\newcommand{\Gcomp}[1][]{
	\,{\mathrel{\mathsmaller{\circ}}}\,
	}                      % Group binary/composition operator 

\newcommand{\homomorphismDiag}[5]{
	\xymatrix{
		#1 \ar@{-}[r]^{#3}    \ar[d]^{#5} & #1 \ar[d]^{#5} \\
		#2 \ar@{-}[r]^{#4}                   & #2
	}
}
\newcommand{\invariantDiag}[5]{
	\xymatrix{
		#1 \ar@{-}[r]^{#3} \ar[dr]_{#4} & #1 \ar[d]^{#4} \\
		& #2 \ar@(r,d)[]^{#5}
	}
}

\newcommand{\isomorphismDiag}[5]{
\xymatrix{
#1 \ar@{-}[r]^{#3}    \ar@<0.5ex>[d]^{#5} \ar@{<-}[d]_{#5^{-1}} & #1 \ar@<0.5ex>[d]^{#5} \ar@{<-}[d]_{#5^{-1}} \\
#2 \ar@{-}[r]^{#4}                   & #2
}
}

\newcommand{\stateSpace}{\mathcal{S}}               % State space
\newcommand{\actionSpace}{\mathcal{A}}              % Action space
\newcommand{\obsSpace}{\mathcal{O}}                      % Observation space
\newcommand{\obsHistorySpace}{\mathcal{X}}             % Observation history space
\newcommand{\obsKernel}{\phi}                      % Observation kernel
\newcommand{\transitionKernel}{\tau}         % Transition kernel
\newcommand{\rewardFn}{r}                           % Reward function
\newcommand{\state}{\vs}
\newcommand{\action}{\va}
\newcommand{\obs}{\vo}
\newcommand{\policy}{\pi}
\newcommand{\expPolicy}{\policy_{\text{exp}}}
\newcommand{\learnedPolicy}{\policy_{\vtheta}}
\newcommand{\obsHistory}[1][t]{\vx_{#1}}
\newcommand{\obsHistoryHorizon}{H}
\newcommand{\dataset}{\mathbb{D}}
\newcommand{\obsHistDataset}{\mathbb{D}^{\text{exp}}_{\obsHistoryHorizon}}
\newcommand{\obsHistAugDataset}{\mathbb{D}^{\text{aug}}_{\obsHistoryHorizon}}

\renewcommand{\KL}{D_{\mathrm{KL}}}
\newcommand{\initDist}{\mathbb{P}_0}

\newcommand{\velocityField}{\vv}
\newcommand{\learnedVelocityField}{\velocityField_{\vtheta}}
\newcommand{\targetVelocityField}{\velocityField_*}

\newcommand{\transportTime}{k}
\newcommand{\transportDist}{\bar{\policy}}

\newcommand{\Unif}{U}

\newcommand{\CE}{\oE_{\scalebox{0.65}{$\rvy \vert \rvx$}}}      % Conditional expectation operator L2 to  L2
\newcommand{\pmd}{\kappa}                    % Kernel of the conditional expectation operator
\newcommand{\mux}{{P_{\rvx}}}    % Measure on X
\newcommand{\muy}{{P_{\rvy}}}    % Measure on Y
\newcommand{\muxy}{{P_{\rvx\rvy}}}    % Measure on X x Y
\newcommand{\nnParams}{{\boldsymbol{\theta}}}
\newcommand{\loss}{\mathcal{L}}
\newcommand{\regularization}{\mathcal{R}}

\let\oldforall\forall
\renewcommand{\forall}{\oldforall \; }
\let\oldexist\exists
\renewcommand{\exists}{\oldexist \: }

\usepackage[nonumberlist,numberedsection]{glossaries}  % for acronyms
\newacronym[
	description={
			Equivariant Neural Conditional Probability: Our proposed model integrating the symmetry priors \cref{eq:main_assumtions} into the NCP deep representation learning algorithm
		}
]{encp}{eNCP}{Equivariant Neural Conditional Probability}
\newacronym[
	description={
			Contrastive Low-Rank loss from \citep{kostic2024neural,ryu2024operator} for operator and representation learning. Used in density-ratio fitting \citep{sugiyama2012density}, representation learning \citep{wang2022spectral,chen2021provable}, and mutual information estimation \citep{tsai2020neuralPMD}
		}
]{clora}{cLoRa}{contrastive low-rank}
\newacronym[
	description={
			Neural Conditional Probability: A deep representation learning framework \citep{kostic2024neural} for conditional probability estimation and regression with statistical guarantees via operator theory \citep{Baker1973-ns}. This framework is symmetry-agnostic
		}
]{ncp}{NCP}{Neural Conditional Probability}
\newacronym[
	description={
			Density Ratio Fitting \citep{tsai2020neuralPMD}: A density ratio \gls{nn} architecture that parameterizes the approximated \gls{pmd} $\pmd_\nnParams: \vsX \times \vsY \to \R_{+}$ as a single \gls{nn}. Consequently, this model cannot be used for downstream conditional probability estimation and regression—it is limited to estimating the mutual information between $\rvx$ and $\rvy$ \citep{tsai2020neuralPMD}
		}
]{drf}{DRF}{Density Ratio Fitting}

\newacronym[
	description={
			Invariant Density Ratio Fitting: This is a $\G$-invariant adaptation of the DRF model \citep{tsai2020neuralPMD} that parameterizes the approximated PMD $\pmd_\nnParams$ as a $\G$-invariant \gls{nn}
		}
]{idrf}{iDRF}{Invariant Density Ratio Fitting}

\newacronym[
	description={Geometric Deep Learning: A field of machine learning that incorporates geometric priors into deep learning models \citep{bronstein2021geometric}}
]{gdl}{GDL}{Geometric Deep Learning}

\newacronym[
	description={Deep Learning}
]{dl}{DL}{Deep Learning}

\newacronym[
	description={Partially Observable Markov Decision Process},
	plural={POMDPs},
	firstplural={Partially Observable Markov Decision Processes}
]{pomdp}{POMDP}{Partially Observable Markov Decision Process}

\newacronym[
	description={Topological Deep Learning. \citep{hajij2022topological}}
]{tdl}{TDL}{Topological Deep Learning}

\newacronym[
	description={Physics-informed ML. \citep{hajij2022topological}}
]{piml}{PiML}{Physics-informed ML}

\newacronym[
	shortplural = PDEs,
	longplural  = Partial Differential Equations,
	description={Partial Differential Equation}
]{pde}{PDE}{Partial Differential Equation}

\newacronym[
	description={Geometric Learning}
]{gl}{GL}{Geometric Learning}

\newacronym[
	shortplural = MSs,
	longplural = Morphological Symmetries,
	description={Morphological Symmetries}
]{ms}{MS}{Morphological Symmetry}

\newacronym[
	description={Robot Geometric Intelligence}
]{rogi}{RoGI}{Robot Geometric Intelligence}

\newacronym[
	description={Machine Learning}
]{ml}{ML}{Machine Learning}

\newacronym[
	description={Reinforcement Learning}
]{rl}{RL}{Reinforcement Learning}

\newacronym[
	description={Behavior Cloning}
]{bc}{BC}{Behavior Cloning}

\newacronym[
	description={
			Pointwise Mutual Dependency \citep{tsai2020neuralPMD}: A pointwise dependency measure between random variables $\rvx$ and $\rvy$, defined as $\pmd(\vx,\vy)=\frac{d\muxy(\vx,\vy)}{d\big(\mux(\vx)\times\muy(\vy)\big)}=\exp\big(\text{MI}(\vx,\vy)\big)$
		}
]{pmd}{PMD}{Pointwise Mutual Dependency}

\newacronym[
	description={
			Conditional Quantile Regression \citep{feldman2023calibrated}: A multivariate neural network approach for regressing upper and lower quantiles at a specified miscoverage level $\alpha$ using pinball loss. Confidence intervals are typically calibrated post-training via conformal prediction, but calibration is omitted here for fair model comparison
		}
]{cqr}{CQR}{Conditional Quantile Regression}
\newacronym[
	description={
			Version of eCQR where the upper and lower parametric quantile functions are parameterized by $\G$-equivariant NNs
		}
]{ecqr}{eCQR}{Equivariant CQR}

\newacronym{ccdf}{CCDF}{Conditional Cumulative Distribution Function}
\newacronym{mlp}{MLP}{Multi-Layer Perceptron}
\newacronym{emlp}{eMLP}{Equivariant MLP}
\newacronym[
	description={
			Conditional Gaussian Mixture Model \citep{gilardi2002conditional}: A parametric model for benchmark conditional density estimation datasets. Generates random variables $\rvx$ and $\rvy$ of arbitrary dimensions with varying mutual information. Enables analytical computation of the \gls{pmd} density ratio (see \cref{sec:background}), unavailable in real-world datasets, allowing direct quantification of approximation error for the conditional expectation operator $\CE$ and its \gls{pmd} density ratio
		}
]{cgmm}{cGMM}{Conditional Gaussian Mixture Model}
\newacronym{mse}{MSE}{Mean Squared Error}
\newacronym{com}{CoM}{Center of Mass}
\newacronym{grf}{GRF}{Ground Reaction Forces}
\newacronym{svd}{SVD}{Singular Value Decomposition}

\newacronym[
	shortplural = NNs,
	longplural  = Neural Networks,
	first={Neural Network (NN)},
	firstplural={Neural Networks (NNs)},
]{nn}{NN}{Neural Network}
\newacronym[
	shortplural = GNNs,
	longplural  = Graph \acrshortpl{nn}s,
	first={Graph \acrshort{nn} (GNN)},
	firstplural={Graph \acrshortpl{nn} (GNNs)},
]{gnn}{GNN}{Graph \acrshort{nn}}
\newacronym[
	shortplural = CNNs,
	longplural  = Convolutional \acrshortpl{nn}s,
	first={Convolutional \acrshort{nn} (CNN)},
	firstplural={Convolutional \acrshortpl{nn} (CNNs)},
]{cnn}{CNN}{Convolutional \acrshort{nn}}

\newacronym[
	shortplural = DoF,                % plural of the short form
	longplural  = degrees of freedom,  % plural of the long form
	first={degree of freedom (DoF)},   % how the first *singular* appearance looks
	firstplural={degrees of freedom (DoF)}, % first *plural* appearance
]{dof}{DoF}{degree of freedom}

\newacronym{ode}{ODE}{ordinary differential equation}

\newacronym{eom}{EoM}{equations of motion}

\newacronym{moma}{MoMa}{Mobile Manipulation}
\newacronym{il}{IL}{Imitation Learning}
\newacronym{fm}{FM}{Flow Matching}
\newacronym{wbc}{WBC}{Whole-Body Controller}
\newacronym{elinear}{eLinear}{equivariant linear}
\newacronym{sota}{s.o.t.a.}{state-of-the-art}
\newacronym{ermsnorm}{eRMSNorm}{equivariant RMS Normalization}
\newacronym{etransformer}{eTransformer}{equivariant transformer}

\makeglossaries

\usepackage{mathtools,amsthm}               % Enhancements for amsmath and theorems
\newtheoremstyle{customplain}%
{0.5em}%          % Space above (preskip)
{0.0em}%          % Space below (postskip)
{\itshape}%                         % Body font (italic for plain style)
{}%                                 % Indent amount (empty = no indent)
{\bfseries}%                        % Theorem head font (bold)
{.}%                                % Punctuation after theorem head
{.5em}%                             % Space after theorem head
{}%                                 % Theorem head spec (empty = normal)
\newtheoremstyle{customremark}%
{6pt plus 2pt minus 1pt}%           % Space above (preskip)
{6pt plus 2pt minus 1pt}%           % Space below (postskip)
{\normalfont}%                      % Body font (normal for remark style)
{}%                                 % Indent amount (empty = no indent)
{\bfseries}%                        % Theorem head font (bold)
{.}%                                % Punctuation after theorem head
{.5em}%                             % Space after theorem head
{}%                                 % Theorem head spec (empty = normal)

\crefname{definition}{Def}{Defs}
\Crefname{definition}{Def}{Defs}
\crefname{section}{Sec.}{Secs.}
\Crefname{section}{Sec.}{Secs.}
\crefname{appendix}{App.}{Apps.}
\Crefname{appendix}{App.}{Apps.}

\theoremstyle{customplain}

\theoremstyle{customremark}

\newtcolorbox[auto counter, number within=section]{hypothesisbox}{
  colback=blue!5,
  colframe=blue!60,
  fonttitle=\bfseries,
  title=Hypothesis~\thetcbcounter,
  sharp corners
}

\usepackage{xcolor}
\definecolor{emerald}{rgb}{0.31, 0.78, 0.47}
\definecolor{linkcyan}{rgb}{0.0, 0.35, 0.35}
\hypersetup{
  colorlinks=true,
  linkcolor=linkcyan,
  citecolor=linkcyan,
  urlcolor=linkcyan,
  filecolor=linkcyan,
  pdfborder={0 0 0}
}
\title{\LARGE \bf Morphologically Equivariant Flow Matching for Bimanual Mobile Manipulation}

\author{Max Siebenborn$^{*1}$, Daniel Ordoñez Apraez$^{*2}$, Sophie Lueth$^1$, \\
Giulio Turrisi$^2$, Massimiliano Pontil$^2$, Claudio Semini$^2$, Georgia Chalvatzaki$^{1,3,4}$
\thanks{$^{1}$ PEARL Lab, Dept. of Computer Science, TU Darmstadt, Germany, $^{2}$Istituto Italiano di Tecnologia, $^{3}$Hessian.AI, $^{4}$Robotics Institute Germany}%
\thanks{$^{*}$Equal contribution.}%
\thanks{This work is supported by ERC grant SIREN (101163933).
Funded by the European Union. Views and opinions expressed are however those of the author(s) only and do not necessarily reflect those of the European Union or the European Research Council Executive Agency. Neither the European Union nor the granting authority can be held responsible for them.}
}

\begin{document}
\maketitle

\begin{strip}
  \vspace{-2.25cm}
  \centering
  \includegraphics[width=\textwidth]{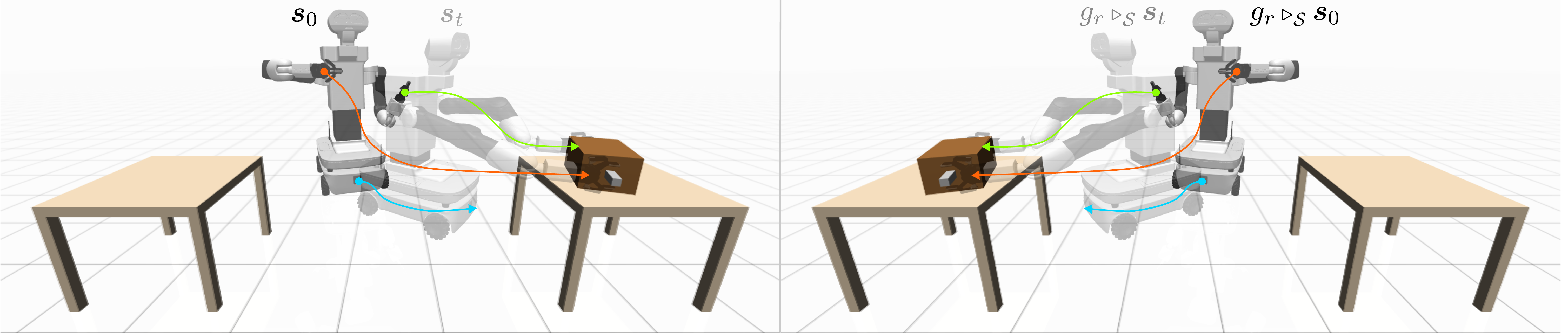}
  \captionof{figure}{Simulated bimanual \textit{box lifting} task, illustrating reflection morphological symmetry of mobile manipulators. 
  A successful trajectory $(\state_t, \action_t)_{t=0..T}$ (\textit{left}) transfers zero-shot to the mirrored setting:
  starting from the reflected initial state $\g_r \Glact[\stateSpace]\state_0$, executing the reflected action sequence $(\g_r \Glact[\actionSpace]\action_t)_{t=0..T}$ produces the trajectory $(\g_r\Glact[\stateSpace]\state_t, \g_r\Glact[\actionSpace]\action_t)_{t=0..T}$ that solves the task in the mirrored setting (\textit{right}).  
  In this paper, we leverage this reflection symmetry prior for behavior cloning with flow matching, demonstrating improved sample efficiency, generalization, and optimality in bimanual mobile manipulation.}
  \label{fig:hero}
\end{strip}

\begin{abstract}
\acrlong{moma} requires coordinated control of high-dimensional, bimanual robots. \acrlong{il} methods have been broadly used to solve these robotic tasks, yet typically ignore the bilateral morphological symmetry inherent in such systems. We argue that morphological symmetry is an underexplored but crucial inductive bias for learning in bimanual \acrlong{moma}: knowing how to solve a task in one configuration directly determines how to solve its mirrored counterpart. In this paper, we formalize this symmetry prior and show that it constrains optimal bimanual policies to be ambidextrous and equivariant under reflections across the robot's sagittal plane. We introduce a $\CyclicGroup[2]$-equivariant \acrlong{fm} policy that enforces reflective symmetry either via a regularized training loss or an equivariant velocity network.
Across planar and 6-DoF \acrlong{moma} tasks, symmetry-informed policies consistently improve sample efficiency and achieve zero-shot generalization to mirrored configurations absent from the training distribution. We further validate this zero-shot generalization capability on a real-world manipulation task with a TIAGo++ robot. Together, our findings establish morphological symmetry as an effective, generalizable, and scalable 
inductive bias for ambidextrous generative policy learning.
    
\end{abstract}

\IEEEpeerreviewmaketitle

% ---------------------------------------
% INTRODUCTION
\section{Introduction}
\let\oldtwocolumn\twocolumn
\noindent
Over the last decade, deep learning has driven remarkable progress in robotics and promises to solve long-standing challenges in service robotics, including household assistance, healthcare support, and collaborative assembly.
Such applications require robots to solve \gls{moma} tasks involving dextrous bimanual manipulation.
Current approaches often rely on \gls{il}, using datasets of expert human demonstrations to train stochastic policies via \gls{bc} \cite{chi2023diffusionpolicy, Zhao2023LearningFB, brohan2022rt, black2024pi_0}. 
However, these methods typically require vast amounts of data and compute due to task complexity and low sample efficiency of generative \gls{il} paradigms. Given the cost and difficulty of collecting expert demonstrations, and the risks of malfunction, 
methods with improved sample efficiency, generalization, and optimality are critically needed.

This paper leverages a core physics-informed inductive bias in bimanual \gls{moma} frequently ignored in \gls{il}: the robot's bilateral morphological symmetry \cite{ordonez2025morphosymm}, rooted in the reflection symmetry of the left and right sides of the body morphologies of humans and most bimanual manipulators (see \cref{fig:hero}).

\begin{figure}
    \centering
    \includegraphics[width=.98\linewidth]{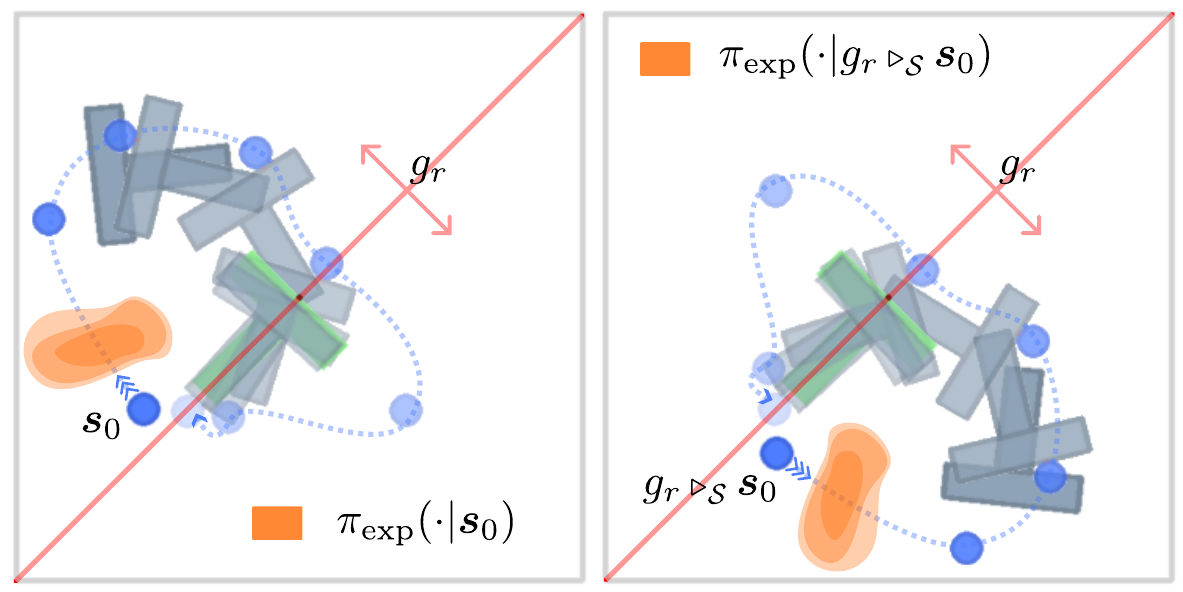}
    \caption{
        Reflection symmetry of the Push-T environment \cite{chi2023diffusionpolicy}, illustrating a rollout starting from state $\state_0$ (left) and the corresponding reflected rollout starting from the reflected state $\g_r \Glact[\stateSpace] \state_0$ (right).
        The expert policy distribution (orange) is reflection-invariant, motivating the use of this symmetry as an inductive bias for policy learning.        
    }
    \vspace*{-0.5cm}
    \label{fig:Push-T_symmetries}
\end{figure}

Intuitively, this symmetry prior implies that observations and actions 
at any time are related by known reflection transformations to those of the mirrored manipulation task (see \cref{fig:hero}). 
Hence, any optimal policy solving one task transfers directly to the reflected task, yielding an \emph{ambidextrous} manipulation policy \cite{li2025morphologically}.
\Cref{fig:Push-T_symmetries} illustrates an analogous control symmetry in the planar Push-T environment \cite{chi2023diffusionpolicy}:
any expert demonstration can be reflected across the vertical symmetry axis of the target-T to produce an expert rollout that solves the task from the reflected initial state.

In this paper, we formalize this symmetry prior for bimanual \gls{moma}, 
reformulating \gls{il} as a symmetry-constrained optimization problem enforceable explicitly or via regularization within \gls{fm}.
Our contributions are:

\begin{itemize}
    \item[(i)] we formalize reflection symmetry-induced equivariance in \gls{fm} for \gls{il} and show two strategies to enforce these constraints: explicitly via equivariant \glspl{nn}, or softly via regularization;
    
    \item[(ii)] we introduce a $\G$-equivariant transformer for arbitrary finite groups, enabling ambidextrous bimanual control;
    
    \item[(iii)] we show that exploiting symmetry improves sample efficiency, generalization, and enables zero-shot transfer to mirrored tasks outside the training distribution.
\end{itemize}

% ---------------------------------------
% METHOD

\section{Problem Formulation}
We model each bimanual manipulation task as a \gls{pomdp} defined by 
$(\stateSpace, \actionSpace, \transitionKernel, \rewardFn, \initDist, \obsSpace, \obsKernel)$
, where $\state \in \stateSpace$ is the world state, $\action \in \actionSpace$ the agent's action, $\transitionKernel:\stateSpace \times \actionSpace \times \stateSpace \rightarrow \R_{+}$ the transition kernel, $\rewardFn: \stateSpace \times \actionSpace \rightarrow \R$ the reward function, $\initDist: \stateSpace \rightarrow \R_{+}$ the initial state distribution, and $o \in \obsSpace$ an observation with $\obsKernel: \obsSpace \times \stateSpace \rightarrow \R_{+}$ the observation model.
We aim to learn a control policy $\learnedPolicy$ from a dataset of $N$ expert trajectories, $\dataset^\text{exp} = \{(\vo^{(n)}_{t}, \va^{(n)}_{t})_{t=0}^{T_n}\}_{n=1}^{N}$, consisting of observation--action sequences generated by an expert policy $\expPolicy$.
We thus frame policy learning as a \acrlong{bc} problem to learn a parametric stochastic policy $\learnedPolicy(\cdot \vert \obsHistory[])$ with parameters $\vtheta$ that approximates the expert policy $\expPolicy(\cdot \vert \obsHistory[])$, where $\obsHistory[]$ a history of $\obsHistoryHorizon$ consecutive observations, $\obsHistory := [\vo_{t-\obsHistoryHorizon+1}, \ldots ,\vo_t] \in \obsHistorySpace \subseteq \obsSpace^{\obsHistoryHorizon}$.
Formally, this objective minimizes the Kullback--Leibler divergence between the expert and learned policy:
    \begin{equation}
        \label{eq:behavior_cloning_objective_kl_form}
        \small
        \nnParams^* = \argmin_{\nnParams} \underset{\obsHistory[] \sim \dataset^{\text{exp}}_{\obsHistoryHorizon}}{\E}
        \KL \left( \expPolicy(\cdot \vert \obsHistory[]) \,\|\, \learnedPolicy(\cdot \vert \obsHistory[]) \right).
    \end{equation}

Crucially, we observe that in bimanual manipulation the \gls{pomdp} admits a unique reflection symmetry, denoted $\g_r$, induced by the robot’s bilateral morphological symmetry \cite{ordonez2025morphosymm} (see \cref{fig:hero}).
This symmetry acts \emph{linearly} on the state, action, and observation spaces via group actions $(\Glact[\stateSpace])\,{:}\, \CyclicGroup[2] \,{\times}\, \stateSpace \,{\to}\, \stateSpace$, $(\Glact[\actionSpace]){:}\, \CyclicGroup[2] \,{\times}\, \actionSpace \,{\to}\, \actionSpace$, and $(\Glact[\obsSpace]){:}\, \CyclicGroup[2] \,{\times}\, \obsSpace \,{\to}\, \obsSpace$, such that for any tuple $(\state, \action, \obs) \in \stateSpace \times \actionSpace \times \obsSpace$ the reflected tuple is given by $(\g_r \Glact[\stateSpace]\state, \g_r \Glact[\actionSpace]\action, \g_r \Glact[\obsSpace]\obs) \in \stateSpace \times \actionSpace \times \obsSpace$. 
Here, $\CyclicGroup[2] := \{e, g_r \mid \g_r^2 = e\}$ denotes the reflection symmetry group, consisting of the identity $e$ and reflection transformation $g_r$.

Intuitively, the reflective symmetry of the \gls{pomdp} implies that under reflected observations, the respective optimal control action should also be reflected.
Formally, the symmetric \gls{pomdp} imposes a $\CyclicGroup[2]$-invariance constraint on the learned stochastic policy \cite{zinkevich2001,li2025morphologically,brehmer2023edgi}:

\begin{subequations}
    \small
    \label{eq:behavior_cloning_objective_kl_equiv_form}
    \begin{align}
        \nnParams^* {=} & \argmin_{\nnParams} \underset{
            \obsHistory[] \sim \dataset^{\text{exp}}_{\obsHistoryHorizon}
        }{\E}
        \KL \left( \expPolicy(\cdot \vert \obsHistory[]) \; \,\|\, \; \learnedPolicy(\cdot \vert \obsHistory[]) \right),
        \\
        \text{s.t.}     \;\;
                        &
        \learnedPolicy(\action \vert \obsHistory[])
        =
        \learnedPolicy(\g_r \Glact[\actionSpace] \action \vert
        \g_r \Glact[\obsHistorySpace] \obsHistory[]),
        \;\;
        \forall\,
        \action \in \actionSpace, \obsHistory[] \in \obsHistorySpace.
        \label{eq:behavior_cloning_objective_kl_equiv_form_invariance_constraint}
    \end{align}
\end{subequations}

Here, $g \Glact[\obsHistorySpace] \obsHistory[] := [\g \Glact[\obsSpace] \obs_t]_{t=1}^{\obsHistoryHorizon}$ denotes the symmetry transformation of an observation-history sample.

\section{Methodology}\label{sec:ws_method}
We address the constrained \acrlong{bc} objective in \cref{eq:behavior_cloning_objective_kl_equiv_form} through \emph{\acrlong{fm}} \cite{lipman2022flow}, where sampling from the policy is modeled as integrating a continuous-time ODE:
\begin{equation}
    \small
    \label{eq:flow_matching_ode}
    \action = \action^{(0)} + \int_{0}^{1} \targetVelocityField(\action^{(\transportTime)}, \obsHistory[], \transportTime) \, \mathrm{d}\transportTime,
    \;\;
    \text{with}\; \action^{(0)} \sim \transportDist_{0}(\cdot \vert \obsHistory[]).
\end{equation}
Here, $\targetVelocityField$ is the velocity field that governs the evolution of the probability path from initial Gaussian noise $\transportDist_{0}=\mathcal{N}(0, I_d)$ to the target expert action distribution $\transportDist_{1}=\expPolicy$.

Previous work \cite{kohler2020equivariant, klein2023equivariant} demonstrates that $\transportDist_{1}$ is $\G$-invariant when both a $\G$-invariant prior (e.g., Gaussian noise) and a $\G$-equivariant velocity field are employed.
Therefore, to address the symmetry-constrained \gls{bc} objective in \cref{eq:behavior_cloning_objective_kl_equiv_form}, we solve a \gls{fm} problem with equivariance constraints:

\begin{subequations}
    \small
    \label{eq:equivariant_fm_optimization}
    \begin{align}
        \argmin_{\vtheta}
        \, &
        \E_{\substack{
        \va^{(\transportTime)} \sim \transportDist_{\transportTime}(\cdot \vert \obsHistory[])
        \\
        \obsHistory[] \sim \obsHistDataset,
        \transportTime \sim \Unif[0,1]
        }}
        \bigl\|
        \learnedVelocityField(\va^{(\transportTime)}, \obsHistory[], \transportTime)
        -
        \targetVelocityField(\va^{(\transportTime)}, \obsHistory[], \transportTime)
        \bigr\|_2^2,
        \label{eq:equivariant_fm_objective}
        \\
        \text{s.t.}\quad
        \; &
        \g_r \Glact[\actionSpace] \learnedVelocityField(\action, \obsHistory[], \transportTime)
        =
        \learnedVelocityField(\g_r \Glact[\actionSpace] \action,
        \g_r \Glact[\obsHistorySpace] \obsHistory[],
        \transportTime),
        \label{eq:equivariant_fm_objective_equiv_constraint}
        \\
           & \qquad\forall \action \in \actionSpace,\; \obsHistory[] \in \obsHistorySpace,\; \transportTime \in [0,1].
        \nonumber
    \end{align}
\end{subequations}
\noindent
While the velocity field error in \cref{eq:equivariant_fm_objective} effectively minimizes the \gls{bc} objective in \cref{eq:behavior_cloning_objective_kl_form} \cite{su2025flow_kl}, enforcing equivariance of the velocity field $\learnedVelocityField$ in \cref{eq:equivariant_fm_objective_equiv_constraint} ensures that the policy $\learnedPolicy$ satisfies the invariance constraint in \cref{eq:behavior_cloning_objective_kl_equiv_form_invariance_constraint} \cite{klein2023equivariant}.

\begin{figure*}[!htb]
    \centering

    \includegraphics[width=\textwidth]{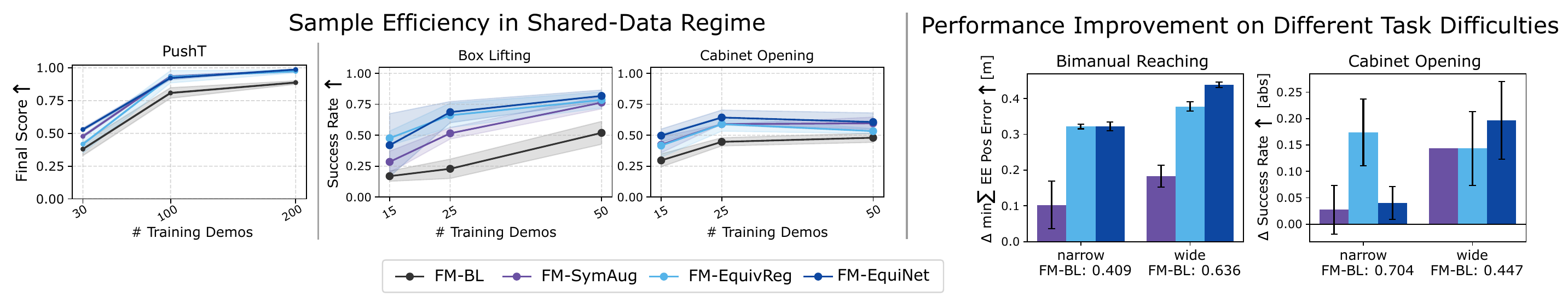}

    \caption{Simulated mobile manipulation results in \textit{shared-data} regime. \textit{Left}: Sample efficiency across simulated tasks — symmetry-informed policies (\emph{SymAug}, \emph{EquivReg}, \emph{EquivNet}) yield on average a
2× gain over the symmetry-agnostic baseline (\emph{BL}). \textit{Right}: Performance improvements of symmetry-informed policies over baseline increase with task difficulty.}
    \label{fig:combined_results}
\end{figure*}

\noindent
In this paper, we consider three ways to leverage the described symmetry prior for policy learning with \gls{fm}:

\begin{itemize}
    \item[(i)] \textbf{Data Augmentation} (\textit{SymAug}, \cref{eq:data_augmentation_dataset}) adds reflected observation-action pairs to the dataset \cite{higgins2022symmetry,ordonez2025morphosymm}:
    \begin{equation}
        \label{eq:data_augmentation_dataset}
        \begin{split}
            \small
            \obsHistAugDataset {:=} \{ (\g \Glact[\obsHistorySpace] \obsHistory[], \g \Glact[\actionSpace] \action) \mid
            (\obsHistory[], \action) \in \obsHistDataset, \g \in \CyclicGroup[2] \}.
        \end{split}
    \end{equation}
    \item[(ii)] The \textbf{Equivariance Consistency Loss} (\textit{EquivReg}, \cref{eq:behavior_cloning_objective_kl_equiv_form_penalty}) encourages reflection symmetry of the velocity field through a regularized training loss, as in \cite{zhang2026equibim}:

\begin{equation}
    \label{eq:behavior_cloning_objective_kl_equiv_form_penalty}
    \argmin_{\vtheta} \, \loss_{\text{CFM}}(\vtheta) + \lambda \underbrace{\left\|
        \substack{
        \g_r \Glact[\actionSpace] \learnedVelocityField(\action^{(\transportTime)}, \obsHistory[], \transportTime)\\
        -\, \learnedVelocityField(\g_r \Glact[\actionSpace] \action^{(\transportTime)},
        \g_r \Glact[\obsHistorySpace] \obsHistory[],
        \transportTime)}
    \right\|^2}_{=:\regularization(\vtheta)}.
\end{equation}

\item[(iii)] The \textbf{Explicit Equivariance Constraint} (\textit{EquivNet}) parameterizes $\learnedVelocityField$ with a $\CyclicGroup[2]$-equivariant \gls{nn}, guaranteeing equivariance also for out-of-distribution inputs. We introduce a novel $\G$-equivariant transformer in which all modules are equivariant to an arbitrary compact group $\G$; details will be released with the final paper.
\end{itemize}

% ------------------

\section{Experiments}

\begin{table}[!t]
	\centering
	\resizebox{.95\linewidth}{!}{%
		\begin{tabular}{l|c|c}
    \multirow{2}{*}{\textbf{Method}} & \textbf{Box Lifting} & \textbf{Cabinet Opening} \\
                                     & \textbf{Zero-shot $e\;$/$\;g_r\;$/total $\uparrow$} & \textbf{Zero-shot $e\;$/$\;g_r\;$/total $\uparrow$} \\
    \midrule
    FM-Baseline  & 0.59 / \textcolor{red}{0.00} / 0.29 & \textcolor{smerald}{0.89} / \textcolor{red}{0.00} / 0.44 \\
    FM-SymAug    & 0.63 / 0.67 / 0.65 & \textcolor{smerald}{0.90} / \textcolor{smerald}{0.88} / \textcolor{smerald}{0.89} \\
    FM-EquivReg  & \textcolor{smerald}{0.70} / \textcolor{smerald}{0.72} / \textcolor{smerald}{0.71} & 0.85 / \textcolor{smerald}{0.87} / 0.86 \\
    FM-EquivNet  & 0.67 / 0.66 / 0.67 & \textcolor{smerald}{0.87} / 0.83 / 0.85 \\
\end{tabular}
}
	\caption{\textit{Zero-shot} evaluation in simulated mobile manipulation: policies are trained on the original configuration ($e$) and tested on both $e$ and its reflection ($g_r$). We report mean success rates over 50 rollouts and three training seeds.}
    \label{tab:zeroshot_results}
\end{table}

Our experiments evaluate the reflective symmetry prior for \gls{il} in terms of sample efficiency, generalization, and optimality.
We compare all symmetry-aware methods from \cref{sec:ws_method} against a symmetry-agnostic \gls{fm} baseline (\emph{BL}) policy.
We consider three settings: (i) the \textit{Push-T} benchmark (\cref{fig:Push-T_symmetries}); (ii) simulated \gls{moma} tasks with the bimanual Tiago++ mobile manipulator (\cref{fig:hero});
and (iii) a real-world Tiago++ task.
We use state-based observations of end-effector and object poses, and end-effector pose control.
For the Tiago++ tasks, all quantities are defined in the robot base frame, and pose targets are tracked by a whole-body controller \cite{Zakka_Mink_Python_inverse_2025, adelprete_jnrh_2016}.

\textbf{Symmetric Zero-Shot Generalization}~~~
\cref{tab:zeroshot_results} demonstrates successful zero-shot transfer of all symmetry-aware methods to mirrored \gls{moma} configurations absent from the training data, matching performance in the original setting.

\textbf{Sample Efficiency}~~~
In a \textit{shared-data} regime, where demonstrations from both the original ($e$) and reflected ($g_r$) tasks are available during training, all symmetry-informed policies consistently outperform the baseline across all dataset size regimes (\cref{fig:combined_results}-\textit{left}).
For the challenging bimanual \textit{box lifting} task we observe additional optimality gains of enforcing the policy invariance optimality constraint \cref{eq:behavior_cloning_objective_kl_equiv_form_invariance_constraint} (\textit{EquivReg}, \textit{EquivNet}) over data augmentation (\textit{SymAug}).

\textbf{Spatial Generalization}~~~
We evaluate two difficulty levels (\textit{narrow} and \textit{wide}) for the \textit{bimanual reaching} and \textit{cabinet opening} tasks, corresponding to narrower and broader distributions of target and cabinet poses. 
\Cref{fig:combined_results}-\textit{right} shows that symmetry-informed policies yield larger performance gains at higher task difficulty, indicating that morphological symmetry priors scale well to more complex tasks.

\begin{figure}[t]
    \centering
    \begin{minipage}{0.51\linewidth}
        \centering
        \includegraphics[width=\linewidth]{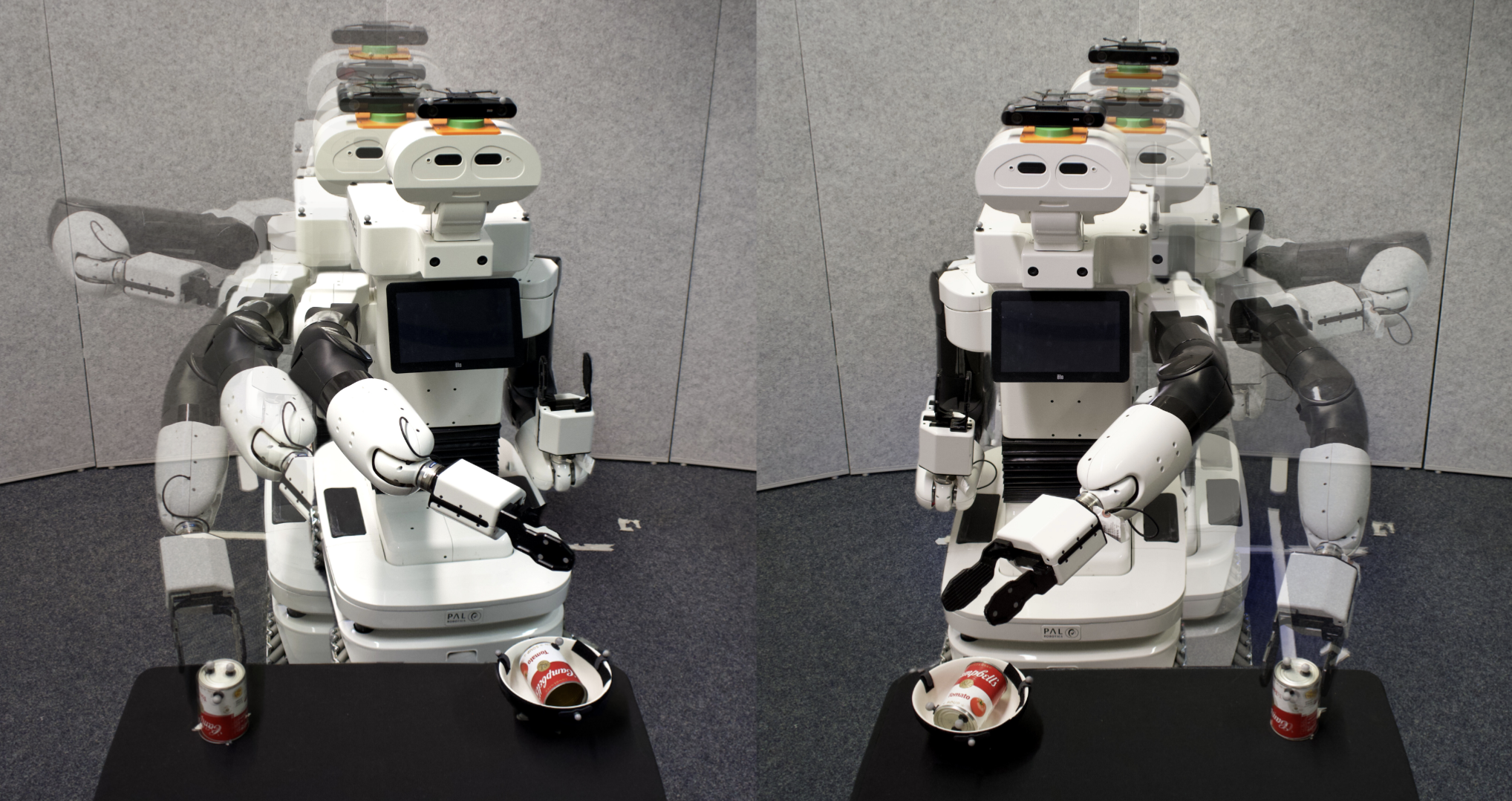}
        \small Mirrored ($g_r$) \hfill Original ($e$)
    \end{minipage}
    \hfill
    \begin{minipage}{0.47\linewidth}
        \centering
        \small
        \begin{tabular}{l|cc}
            \textbf{Method} & $e$ & $g_r$ \\
            \midrule
            Baseline  & 10/10 & \textcolor{red}{0/10} \\
            SymAug    & 10/10 & 10/10 \\
            EquivReg  & 10/10 & 10/10 \\
            EquivNet  & 10/10 & 10/10 \\
        \end{tabular}
    \end{minipage}
    \caption{Real-world Tiago++ results: symmetry-informed policies generalize \textit{zero-shot} to mirrored can-in-bowl task ($g_r$).}
    \label{fig:real_task}
\end{figure}
\textbf{Real-World Results}~~~
Finally, we validate the morphological symmetry prior on a real static-base unimanual Tiago++ task in \textit{zero-shot} setting:
the policy is trained on a pick-and-place task with the left arm, and evaluated both in this original ($e$) and the mirrored setting ($g_r$, right arm) (\cref{fig:real_task}-\textit{left}).
\Cref{fig:real_task}-\textit{right} confirms repeatable zero-shot transfer under real-world conditions, validating the morphological symmetry prior beyond idealized simulation.

\section{Related Work}
Previous work shows successful \gls{il} in \gls{moma} \cite{fu2024mobile, yang2025mobipi, brohan2022rt, shafiullah2023bringing,sundaresan2025homer,bahety2025safemimic}.
Symmetry priors in generative \gls{il} have been used to improve sample efficiency and generalization \cite{wang2024equivariant, wang2025practical, tie2024etseedefficienttrajectorylevelse3, yang2024equibot,yang2024equivact,jankowski2025guaranteed,jia2022seilsimulationaugmentedequivariantimitation, funk2024actionflow, chang2026efficientflow}, mostly focusing on $\SE[3]$-equivariance. 
Morphological symmetry priors have been studied for locomotion \cite{mittal2024symmetry, 2019-MIG-symmetry, su2024leveraging, wei2025ms, xie2024morphologicalsymmetryequivariantheterogeneousgraphneural} and ambidextrous manipulation \cite{li2025morphologically, zhang2026equibim}.
The work most closely related to ours is the concurrent study of \cite{zhang2026equibim}, which exploits morphological symmetry for \gls{il} in bimanual manipulation via regularization. In contrast, our work addresses \gls{moma}, focusing on generative policy learning with \gls{fm}, with symmetry-priors enforced via equivariant \glspl{nn}, loss regularization, and data augmentation.

%%%%% CONCLUSION

\section{Conclusions}
We introduced a \gls{fm} framework for \gls{il} that, by construction, respects the bilateral morphological symmetry of mobile manipulators.
Our results show improved sample efficiency, generalization, and optimality, and indicate that these benefits scale with increasing task complexity.
Further, our analysis highlights the advantages of imposing the symmetry prior as an optimality constraint (\cref{eq:equivariant_fm_objective_equiv_constraint}; \textit{EquivReg}, \textit{EquivNet}) over data augmentation.
In future work, we will integrate visual observations  
and scale our method to temporal and rotational symmetries in the environment.

\newpage
\bibliographystyle{IEEEtran}
\bibliography{bibliography}

\end{document}